\documentclass[
]{ceurart}

\usepackage{listings}

\usepackage{amssymb}
\usepackage{latexsym}
\usepackage{soul}

\usepackage{url}
\usepackage{xcolor}
\definecolor{newcolor}{rgb}{.8,.349,.1}

\definecolor{codegreen}{rgb}{0,0.6,0}
\definecolor{codegray}{rgb}{0.5,0.5,0.5}
\definecolor{codepurple}{rgb}{0.58,0,0.82}
\definecolor{backcolour}{rgb}{1.0,1.0,1.0}

\lstdefinestyle{mystyle}{
	backgroundcolor=\color{backcolour},   
	commentstyle=\color{codegreen},
	keywordstyle=\color{magenta},
	numberstyle=\tiny\color{codegray},
	stringstyle=\color{codepurple},
	basicstyle=\ttfamily\footnotesize,
	breakatwhitespace=false,         
	breaklines=true,                 
	captionpos=b,                    
	keepspaces=true,                 
	numbers=left,                    
	numbersep=5pt,                  
	showspaces=false,                
	showstringspaces=false,
	showtabs=false,                  
	tabsize=2
}

\usepackage{booktabs}%
\usepackage{algorithm}%
\usepackage{algorithmicx}%
\usepackage{algpseudocode}%
\usepackage{amsmath,amssymb,amsfonts}%
\usepackage{graphics}
\usepackage{todonotes}
\usepackage{pifont}

\usepackage{listings}

\newcommand{\cmark}{\ding{51}}%
\newcommand{\xmark}{\ding{55}}%
\begin{document}
	
	\copyrightyear{2024}
	\copyrightclause{Copyright for this paper by its authors.
		Use permitted under Creative Commons License Attribution 4.0
		International (CC BY 4.0).}
	
	\conference{Accepted at AEQUITAS 2024: Workshop on Fairness and Bias in AI | co-located with ECAI 2024, Santiago de Compostela, Spain}
	
	\title{FairX: A comprehensive benchmarking tool for model analysis using fairness, utility, and explainability}
	
	
	\author[1]{Md Fahim Sikder}[%
	orcid=0000-0001-5307-997X,
	email=md.fahim.sikder@liu.se,
	url=https://fahimsikder.com/,
	]
	\cormark[1]
	\address[1]{Department of Computer and Information Science (IDA), Link\"oping University, Sweden}
	
	\author[2]{Resmi Ramachandranpillai}[
	email=r.ramachandranpillai@northeastern.edu
	]
	
	\address[2]{Institute for Experiential AI, Northeastern University, USA}
	
	\author[1]{Daniel {de Leng}}[%
	email=daniel.de.leng@liu.se
	]
	
	\author[1]{Fredrik Heintz}[%
	email=fredrik.heintz@liu.se
	]

	\cortext[1]{Corresponding author.}
	
	\begin{abstract}
		We present FairX, an open-source Python-based benchmarking tool designed for the comprehensive analysis of models under the umbrella of fairness, utility, and eXplainability (XAI). FairX enables users to train benchmarking bias-mitigation models and evaluate their fairness using a wide array of fairness metrics, data utility metrics, and generate explanations for model predictions, all within a unified framework. Existing benchmarking tools do not have the way to evaluate synthetic data generated from fair generative models, also they do not have the support for training fair generative models either. In FairX, we add fair generative models in the collection of our fair-model library (pre-processing, in-processing, post-processing) and evaluation metrics for evaluating the quality of synthetic fair data. This version of FairX supports both tabular and image datasets. It also allows users to provide their own custom datasets. The open-source FairX benchmarking package is publicly available at \url{https://github.com/fahim-sikder/FairX}.
	\end{abstract}
	
	\begin{keywords}
		Fair evaluation \sep
		Benchmarking tool \sep
		Synthetic data \sep
		Data utility \sep
		Explainability
	\end{keywords}
	
	\maketitle
	
	\section{Introduction}
	
	
	
	
	

	With the rapid development of artificial intelligence-based systems to aid us in our daily lives, it is important for these systems to give outcomes that is acceptable for all users, including---but not limited to---from demographic perspective. 
	Troublingly, as the available data is filled with human or machine bias, models trained with these dataset often gives unfair outcome towards some demographic~\cite{liu2022fair}.
	It is therefore critical to mitigate bias in the dataset and model.
	Over the years, researchers have used different techniques to achieve this~\cite{ntoutsi2020bias, mehrabi2021survey}. 
	These techniques can be roughly grouped into three families: 1) \emph{Pre-processing}, i.e.~where the dataset is processed in such a manner that it produces less biased outcomes, before passing it to a model for training; 2) \emph{In-processing}, i.e.~where the model learns the original data distribution and shifts the data distribution to a fair distribution by adding some constraints during the training process; and 3) \emph{Post-processing}, i.e.~where the model's outcome is changed in such a manner that it gives fair outcomes relative to protected attributes.
	The performance of these models or datasets can be measured by the evaluation metrics that reflect both the fairness and data utility. To ease up the work for training models and evaluating them, researchers has developed benchmarking tool that bring the training and evaluation in one framework. Recently, research on fair generative models has found a lot of spotlight and measuring the quality of the synthetic data is as crucial as evaluating fairness and data utlity.

	

	

	
	Existing fairness-related benchmarking tools focus on creating benchmarks and measuring their fairness on different datasets. 
	For example, FairLearn~\cite{weerts2023fairlearn} by Microsoft contains several fair models and evaluation metrics for checking fairness and data utility. 
	AI Fairness 360 (AIF360)~\cite{bellamy2019ai} by IBM also contains fairness evaluation metrics and basic data utility measuring metrics. 
	But both of these frameworks lack the ability to train fair generative models and measure the data utility for synthetic data. 
	For synthetic fair data, it is important to validate the quality of the generated data alongside measuring the fairness and other data utilities. 
	Explainability is an essential property of fair models because it aids in making the model's decision-making process more transparent.
	These modules should therefore be included in such benchmarking tools. 
	
	\begin{figure}[t]
		\centering
		\includegraphics[width=\textwidth,height=4.5cm]{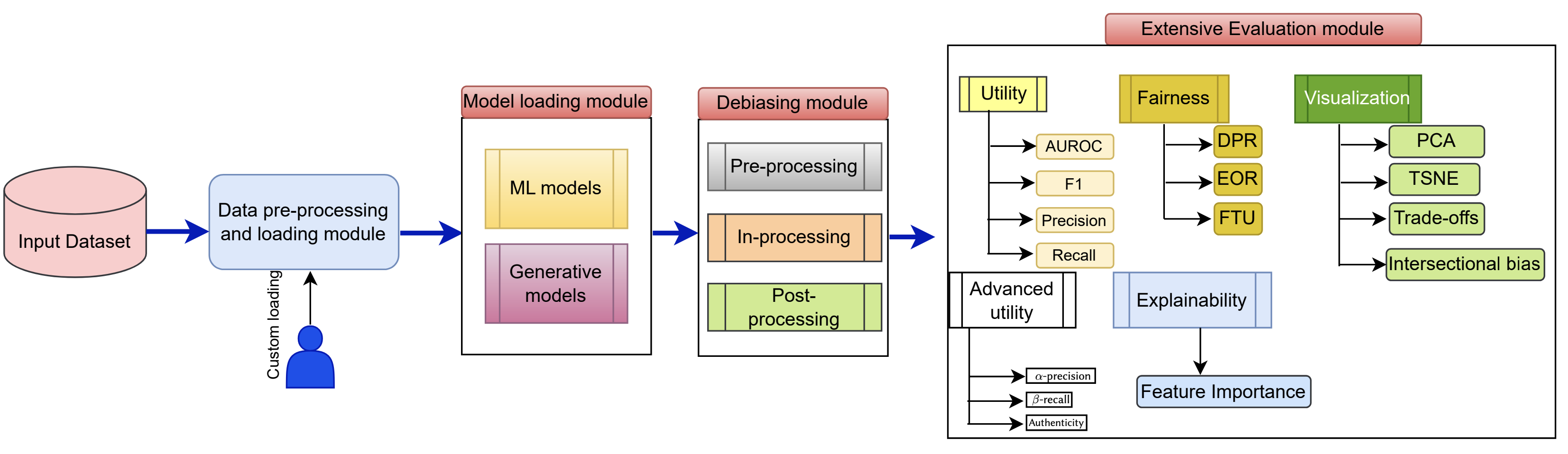}
		\caption{A High-level overview of FairX. An input dataset (possibly custom) is fed to the FairX data loading module followed by a bias-mitigation module and an extensive evaluation module providing multi-faceted evaluations.}
		\label{fig:FairX}
	\end{figure}
	
	\textbf{In this work, we present \emph{FairX}, an open-source modular fairness benchmarking tool, available to use at \url{https://github.com/fahim-sikder/FairX}.}
	A high-level system overview is given in Figure~\ref{fig:FairX}.
	FairX contains data processing techniques and benchmarking fairness models (incorporating pre-processing, in-processing, and post-processing), including generative fair models. 
	We evaluate these models in terms of fairness, data utility. 
	We also add evaluation methods for synthetic fair data (Advanced Utility) to check the quality of the generated samples. 
	FairX supports both tabular and image data and can plot feature importance for down-streaming task using explainable algorithms.
	
	The remainder of this paper is organised as follows.
	In Section~\ref{sec:background} we discuss some background information that will help the reader understand the rest of the paper.
	We then present FairX in Section~\ref{sec:fairx}.
	Section~\ref{sec:results} shows some fairness results obtained by FairX for a number of datasets and models.
	Finally, the paper looks ahead towards future improvements in Section~\ref{sec:conclusion}.

	\section{Background} \label{sec:background}
	
	In this section, we provide the necessary details to follow the paper.
	
	\subsection{Bias mitigation methods}
	
	A variety of bias mitigation methods have been proposed in the literature based on data, training, and predictions. These methods can be broadly categorized into three main approaches: pre-processing, in-processing, and post-processing techniques.
	
	\paragraph{Pre-processing.} These techniques involve altering the training data to resolve any potential causes of biases before it is fed to the model. There are various techniques in the literature such as disparate impact remover \cite{feldman2015certifying}, data cleaning and augmentation, and fair representation learning \cite{zemel2013learning}. This involves balancing the representation of different groups or generating synthetic data to augment underrepresented groups, assigning weights to uphold some minority groups, and transforming the data representation in a format that obscures protected features while maintaining feature attributions. 
	
	\paragraph{In-processing.} This involves mitigating biases during training. The techniques involve fairness constraints, adversarial de-biasing \cite{zhang2018mitigating}, and fairness-aware learning. In fairness constraints training, a multi-objective optimization combining a prediction loss and a fairness penalty will be used such as adding regularization terms to the objective function that penalizes unfairness or incorporating fairness metrics as part of the optimization process. In adversarial de-biasing \cite{zhang2018mitigating}, adversarial training is used to reduce bias. The model is trained to perform well on the primary classification/prediction tasks while simultaneously trying to prevent an adversary from predicting the protected features, thus forcing the model to learn less biased representations.
	
	\paragraph{Post-processing.} These methods are applied to the predictions of a classifier. Techniques such as threshold adjustment, calibration \cite{pleiss2017fairness}, and Reject Option Classifications \cite{kamiran2012decision} fall under this category. In threshold adjustment the decision thresholds of a trained model are adjusted to ensure that the outcomes meet the chosen fairness metric. Calibration \cite{pleiss2017fairness} ensures that the predicted probabilities maintain the true likelihood of outcomes equally across different demographic groups. Techniques like equalized odds post-processing is used where the model's outputs are adjusted to satisfy fairness constraints. Reject Option-Based Classification (ROC) \cite{kamiran2012decision} allows the model to prevent from making a decision when the confidence is low, for the chosen sensitive attributes. This can reduce the likelihood of biased or unfair decisions in uncertain instances.

	\subsection{Evaluation metrics}
	
	To measure the performance of models or dataset, various evaluation methods are being used. For evaluating fair model or checking the dataset for potential bias, different kinds of fairness metrics exists. For example, demographic parity checks if the decision from a down-streaming task is equal for each class in sensitive attributes. Fairness through unawareness \cite{cornacchia2023auditing} checks how the accuracy of down-stream task effects if no-sensitive attributes is used during the training and prediction phase. 
	Adding fairness constraints to the models or datasets may change the data distributions and thereby affect the performance of the dataset or models \cite{wang2022understanding}. 
	To check the data utility performance, we commonly use Accuracy score, F1-score, Precision and Recall. To evaluate the quality of the synthetic data researchers use, $\alpha$-precision \cite{alaa2022faithful}, $\beta$-recall \cite{alaa2022faithful}. Also to check, is the generative model is truly generating new contents or not, the metrics authenticity \cite{alaa2022faithful} is being used.
	
	\subsection{Comparison of existing benchmarking tools}
	
	Over the years researchers have developed various fairness benchmarking tools which commonly include a dataset loader, different bias mitigation techniques and evaluation metrics. Fairlearn~\cite{weerts2023fairlearn} by Microsoft is one such benchmarking tool. It has support for different algorithms for bias mitigation and measuring the fairness of a model. AIF360~\cite{bellamy2019ai} by IBM is another benchmarking tool. It supports a wide range of evaluation metrics (both for fairness and data utility) and bias-removal algorithms (in-processing, pre-processing and post-processing). Another example is Jurity~\cite{thielbar2023surrogate}. It contains recommender system evaluations, and various fairness and data utility functions. AEQUITAS \cite{saleiro2018aequitas}, FairBench \cite{krasanakis2024towards} generate fairnes report and REVISE \cite{revisetooleccv} is a tool to detect and mitigate bias in the image dataset. More recently, in the area of generative models, there has been an increased interest in generating fair data in the image, tabular and medical domains \cite{ramachandranpillai2024bt, li2022fairgan, ramachandranpillai2023fair, liu2022fair, rajabi2022tabfairgan, van2021decaf}. But the aforementioned benchmarking tools do not contain these models. Also, when evaluating models, other benchmarking tools, only measure the fairness and data utility of the models itself. But evaluation methods for generated data is needed. We need to verify the quality of the synthetic data. We also need to verify the authenticity of the synthetic data, to show the generative models are actually generating new content rather than just copying the data itself. FairX is bridging this gap. We add support for evaluating synthetic data and add generative models in our benchmarking tool. Table \ref{tab:comparison-related} shows the comparison among the models with FairX.

	\begin{table*}[t]
		\centering
			\caption{Comparison of existing benchmarking tools with FairX over different key areas of interests: Fairness Evaluation; Synthetic Data Evaluation; Model Explainability; and Generative Fair Model Training.}
			\resizebox{.8\columnwidth}{!}{
				\begin{tabular}{lcccc}\toprule
					\textbf{Benchmarking}     &\textbf{Fairness Evaluation}    &\textbf{Synthetic Data}    &\textbf{Explainability} & \textbf{Generative Model} \\ 
					\textbf{Tools} & & \textbf{Evaluation} & & \textbf{Training} \\
					\midrule
					Fairlearn \cite{weerts2023fairlearn}  &\cmark   & \xmark     & \xmark  &  \xmark\\
					AIF360 \cite{bellamy2019ai} & \cmark & \xmark & \cmark & \xmark\\
					Jurity \cite{thielbar2023surrogate} & \cmark & \xmark & \xmark & \xmark\\
					AEQUITAS \cite{saleiro2018aequitas} & \cmark & \xmark & \xmark & \xmark\\
					REVISE \cite{revisetooleccv} & \cmark & \xmark & \xmark & \xmark\\
					FairBench \cite{krasanakis2024towards} & \cmark & \xmark & \xmark & \xmark\\
					FairX (ours) & \cmark & \cmark & \cmark & \cmark\\
					\bottomrule
				\end{tabular}
			}
			\label{tab:comparison-related}
		\end{table*}

		\section{FairX} \label{sec:fairx}
		
		
		
		In this section we present FairX in detail. FairX is built on three primary modules, 1) the \emph{Data Loading Module}, 2) the \emph{Bias-mitigating Techniques Module}, and 3) the \emph{Evaluation Module}. The main pipeline (shown in Figure~\ref{fig:FairX}) works as follows. Given a dataset, FairX will pre-process it in a way that is compatible with the benchmarking model. Next the model will train itself using the dataset. After the training the evaluation module will give the results based on fairness, data utility and explain the outcome using explainability.
		
		\subsection{Data loading module}
		
		The \textsc{BaseDataClass} handles the internal processing of datasets and make it compatible with the bias-mitigating models that are present on our framework as well as making it easier to handle for other bias-mitigating models that are not present in this tool. This class contains different methods for handling different kinds of data extension (CSV, and others). We add three widely used tabular datasets (Adult-Income, COMPAS and Credit Card) and two image datasets (Colored MNIST and CelebA) in the benchmarking tool, and we plan to add more. The \textsc{BaseDataClass} process datasets based on numerical and categorical features. It also provides methods to normalize the dataset and is equipped with functionality for various encodings (e.g.~One-hot encoding, QuantileTransformer). It also has a dataset-splitting function to split the dataset for training and testing purpose. We also add functionality to prepare the dataset for explainability algorithms. Sample usage of datasets are described in Appendix Section \ref{sub:dataset-usage}, Listing 1.
		
		\paragraph{Custom Dataset Loader.} Besides adding widely used benchmarking datasets for fair data research, we also provide the option to use custom dataset. By using the \textsc{CustomDataClass}, users can load their own dataset (CSV, TXT, etc.) and train the models. Users need to specify the sensitive attributes and target attributes while using the \textsc{CustomDataClass}. Pre-processing and other functionalities are also available in this class, like in the \textsc{BaseDataClass}. We present sample usage of \textsc{CustomDataClass} in Listing 4 of Appendix Section \ref{sub:metrics-usage}.

		
		\subsection{Bias-mitigating techniques module}
		
		One of FairX's main aims is to benchmark different bias-mitigation techniques on various datasets. Over the years, different techniques have been proposed, and we add models from these techniques to the tool. For the benchmarking process, we use the same hyper-parameters used in their respective works. We create a common format for all the bias-mitigation techniques to make it easy for the users. For example, each bias-mitigation technique has its own class, which has \textsc{model.fit()} function. This \textsc{fit()} function takes the dataset and processes it (if needed for the specific model). For the generative models (in-processing techniques), this function also generates synthetic data and saves it as a Pandas dataframe. Sample usage of models is described in Appendix Section \ref{sub:model-usage}, Listing 2.
		
		\paragraph{Pre-processing.} We add the support for Correlation remover~\cite{weerts2023fairlearn} (\textsc{CorrRemover} in FairX) in the benchmarking. Correlation Remover removes the correlation between the sensitive attributes with other data features by using a linear transformation while keeping as much information as possible. It is also possible to control on how much correlation we want remove by using the \textsc{remove\_intensity} parameter while the value $1.0$ will result maximum correlation removal while $0.0$ will do the opposite. We can access the pre-processing algorithm by using \textsc{fairx.models.preprocessing}. 
		
		\paragraph{In-processing.} Most recent in-processing bias mitigation techniques are based on generative models. And the fairness benchmarking tools we mentioned in this work does not contain these models. One of our contribution of FairX is that, we add several fair generative models, such as, TabFairGAN \cite{rajabi2022tabfairgan}, Decaf \cite{van2021decaf} and Fairdisco \cite{liu2022fair}. We can access the in-processing algorithm by using \textsc{fairx.models.inprocessing} module. After training, these models will generate and save the samples automatically.
		
		\paragraph{Post-processing.} For the post-processing bias mitigation technique, We add Threshold Optimizer~\cite{weerts2023fairlearn}. This technique operates on a classifier and improve the output of its based on a fairness constraint. In this case, we use \textsc{demographic\_parity} as a fairness constraint to improve the outcome of the classifier as presented in \cite{weerts2023fairlearn}. For using the post-processing algorithm, we can use \textsc{fairx.models.postprocessing} module.

		\begin{table}[t]
			\caption{Breakdown of FairX-supported features.}
			\resizebox{0.6\columnwidth}{!}{
				\begin{tabular}{@{}lll@{}}
					\toprule
					\textbf{Dataset} &  & \begin{tabular}[c]{@{}l@{}}Adult-Income, Compas, Credit-card\\ Colored MNIST (Image)\\ CelebA (Image)\end{tabular} \\ \midrule
					\multirow{3}{*}{} & Pre-processing & \begin{tabular}[c]{@{}l@{}}Correlation Remover\end{tabular} \\ \cmidrule(l){2-3} 
					\textbf{Models} & In-processing & \begin{tabular}[c]{@{}l@{}}TabFairGAN \cite{rajabi2022tabfairgan}\\ FairDisco \cite{liu2022fair}\\ Decaf \cite{van2021decaf}\end{tabular} \\ \cmidrule(l){2-3} 
					& Post-processing & Threshold Optimizer \\ \midrule
					\multirow{3}{*}{} & Fairness & \begin{tabular}[c]{@{}l@{}}Demographic Parity Ratio (DPR)\\ Equilized Odds Ratio (EOR)\\ Fairness through Unawareness (FTU)\end{tabular} \\ \cmidrule(l){2-3} 
					\textbf{Metrics}  & Data Utility & \begin{tabular}[c]{@{}l@{}}AUROC, F1-score, Precision\\ Recall, Accuracy\end{tabular} \\ \cmidrule(l){2-3} 
					& Synthetic Data Evaluation & \begin{tabular}[c]{@{}l@{}}$\alpha$-precision \cite{alaa2022faithful}\\ $\beta$-recall \cite{alaa2022faithful}\\ Authenticity \cite{alaa2022faithful}\\ \end{tabular} \\ \midrule
					\textbf{Plotting} &  & \begin{tabular}[c]{@{}l@{}} PCA \cite{bryant1995principal} \& t-SNE \cite{van2008visualizing} plots\\ Feature Importance \\ Fairness vs Accuracy \\ Intersectional Bias \end{tabular} \\ \midrule
					\textbf{Explainability} &  & \begin{tabular}[c]{@{}l@{}}Explain prediction of a model\\ Feature Importance\end{tabular} \\ \bottomrule
				\end{tabular}
			}
			\label{tab:table5}
		\end{table}

		\subsection{Evaluation module}
		
		In FairX, we aim to evaluate the performance of model or dataset using wide range of evaluation metrics. We evaluate in terms of fairness, data utility. Other existing fairness benchmarking tools, lacks the capability to measure the data quality of the synthetic data. It is necessary to check the data quality of the synthetic data as well as the fairness criteria. Here, we present the evaluation module FairX has and we use \textit{XGBoost} as a classifier, also we keep the option to use scikit-learn's \textsc{LogisticRegression}.
		
		
		
		
		\paragraph{Fairness Evaluation.} We create the \textsc{FairnessUtils} class to accommodate fairness evaluation metrics. In this class, currently we add the support for checking the Demographic Parity Ratio, Equalized Odds Ratio, Fairness Through Unawareness (FTU) metrics. We also have plan to add more metrics over the time. Fairness metrics can be accessed using the \textsc{fairx.metrics.FairnessUtils} module. 
		
		\paragraph{Data Utility.} Beside checking the fairness criteria of the datasets or models, we also add the functionality to check the data utility using FairX. We add the support for checking the Accuracy, Precision, Recall, AUROC, and F1-score. And these functions can be accessed by using the \textsc{fairx.metrics.DataUtilsMetrics} module.
		
		\paragraph{Synthetic Data Evaluation.} In FairX, we add the functionality to evaluate the quality of the generated data by the fair generative models. It is important to validate the quality of the synthetic data along with the validation of fairness and data utility criteria. Existing fairness measuring benchmark do not have the functionality to evaluate the synthetic data quality. We evaluate the synthetic data quality in terms of fidelity, diversity and check if the synthetic data has any trace of original data in it \cite{sikder2023transfusion}. We use $\alpha$-precision \cite{alaa2022faithful} to evaluate the fidelity of the synthetic data, $\beta$-recall \cite{alaa2022faithful} to check the diversity and Authenticity \cite{alaa2022faithful} is used to check if the generative models are just memorising the training data or not. Synthetic data evaluation module can be accessed from \textsc{fairx.metrics.SyntheticEvaluation}. We also add the t-SNE and PCA plots to check the fidelity and diversity of the synthetic data too, more about the plots are discussed in section \ref{subsec:plot}. 
		
		\paragraph{Explainability.} We add the explainability functionality in FairX to explain the prediction of a model. We train a classifier (XGBoost) on the benchmarking datasets, and then we explain the prediction using the \textsc{fairx.explainability.ExplainUtils} module. This module is based on the \textsc{TreeExplainer} of SHAP \cite{lundberg2020local2global}. Beside this, we give the functionality to show the feature importance while making a decision. This functionality is especially useful when we want to see how much importance is given to the sensitive attributes while making a decision.
		
		\subsection{Plotting}
		\label{subsec:plot}
		
		
		We add various plotting support in FairX. They can be accessed under the \textsc{fairx.utils.plotting} module. We add support to show the performance trade-off of model accuracy and their fairness performance. Also, we plot the feature importance to show which features are responsible for prediction outcome. This comes in handy analyzing original data and synthetic fair data to see how much the fair model reduce the feature importance for the sensitive attributes.
		
		To show the quality of the synthetic data generated by the fair generative models, we add PCA and t-SNE plots. These plots shows how close the synthetic data is from the original data.
		


		

		\begin{table}[t]
			\caption{Evaluation on the \textbf{Adult-Income} dataset using different models presented at the FairX. Bold indicates best result, and all the metrics score are higher as better. Synthetic Data Evaluation is only applicable to the Fair Generative Models (i.e.~TabFairGAN and Decaf).}
			\centering
				\resizebox{1.0\columnwidth}{!}{%
					\begin{tabular}{lccccccccc}\toprule
						& & \multicolumn{2}{c}{\textbf{Fairness Metrics}} &  \multicolumn{3}{c}{\textbf{Data Utility}} & \multicolumn{3}{c}{\textbf{Synthetic Data Evaluation}}\\
						\cmidrule(lr){3-4}  \cmidrule(lr){5-7} \cmidrule(lr){8-10}
						&\textbf{Protected} & DPR & EOR & ACC & AUC & F1- & $\alpha$- & $\beta$- & Authenticity\\ 
						& \textbf{Attribute} & & & & & Score & precision & recall &\\
						
						\midrule
						
						Correlation-  & Gender  & 0.32 $\pm$ .01 & 0.23 $\pm$ .01 & 0.86 $\pm$ .01 & 0.79 $\pm$ .01 & 0.71 $\pm$ .01 & \cellcolor{gray!25}n/a & \cellcolor{gray!25}n/a & \cellcolor{gray!25}n/a\\
						Remover & Race   & 0.29 $\pm$ .01 & 0.20 $\pm$ .01 & 0.86 $\pm$ .01 & \textbf{0.80 $\pm$ .01} & 0.71 $\pm$ .01 & \cellcolor{gray!25}n/a & \cellcolor{gray!25}n/a & \cellcolor{gray!25}n/a \\
						\midrule
						
						
						
						TabFairGAN & Gender  & 0.69 $\pm$ .01& 0.60 $\pm$ .01 & 0.84$\pm$ .01 & 0.76 $\pm$ .01 & 0.65 $\pm$ .01& \textbf{0.91 $\pm$ .01} & \textbf{0.50 $\pm$ .01} & \textbf{0.537 $\pm$ .01} \\
						& Race  & 0.026 $\pm$ .01 & 0.00 $\pm$ .00 & 0.84 $\pm$ .01 & 0.77 $\pm$ .01 & 0.67 $\pm$ .01 & \textbf{0.98 $\pm$ .01} & \textbf{0.489 $\pm$ .01} & \textbf{0.52 $\pm$ .01} \\
						\midrule

						Decaf & Gender  & 0.52 $\pm$ .01& 0.42 $\pm$ .01 & 0.75 $\pm$ .01& 0.63 $\pm$ .01 & 0.44 $\pm$ .01& 0.07 $\pm$ .01 & 0.08 $\pm$ .01 & 0.009 $\pm$ .01\\
						& Race & 0.55 $\pm$ .01 & 0.46 $\pm$ .01 & 0.77 $\pm$ .01 & 0.69 $\pm$ .01 & 0.53 $\pm$ .01 & 0.06 $\pm$ .01 & 0.08 $\pm$ .01 & 0.009 $\pm$ .01\\
						
						\midrule
						
						FairDisco & Gender  & \textbf{0.98 $\pm$ .01} & \textbf{0.85 $\pm$ .01} & 0.78 $\pm$ .01 & 0.63 $\pm$ .01 & \textbf{0.86 $\pm$ .01} & \cellcolor{gray!25}n/a & \cellcolor{gray!25}n/a & \cellcolor{gray!25}n/a \\
						& Race  & \textbf{0.95 $\pm$ .01} & \textbf{0.92 $\pm$ .01} & 0.812 $\pm$ .01 & 0.71 $\pm$ .01 & \textbf{0.88 $\pm$ .01} & \cellcolor{gray!25}n/a & \cellcolor{gray!25}n/a & \cellcolor{gray!25}n/a \\
						\midrule
						
						
						Threshold  & Gender   & 0.95 $\pm$ .01 & 0.35 $\pm$ .01 &0.86 $\pm$ .01 & 0.66 $\pm$ .01 & 0.65 $\pm$ .01  & \cellcolor{gray!25}n/a & \cellcolor{gray!25}n/a & \cellcolor{gray!25}n/a\\
						Optimizer& Race  & 0.69 $\pm$ .01 & 0.25 $\pm$ .01 & 0.87 $\pm$ .01 & 0.66 $\pm$ .01 & 0.71 $\pm$ .01& \cellcolor{gray!25}n/a & \cellcolor{gray!25}n/a & \cellcolor{gray!25}n/a\\
						
						\midrule
						
						Original Data  & Gender  & 0.32 $\pm$ .01 & 0.22 $\pm$ .01& \textbf{0.88 $\pm$ .01} & \textbf{0.80 $\pm$ .01}& 0.72 $\pm$ .01& \cellcolor{gray!25}n/a & \cellcolor{gray!25}n/a & \cellcolor{gray!25}n/a\\
						& Race  & 0.19 $\pm$ .01& 0.00 $\pm$ .00 &\textbf{0.88 $\pm$ .01}  & \textbf{0.80 $\pm$ .01}& 0.71 $\pm$ .01& \cellcolor{gray!25}n/a & \cellcolor{gray!25}n/a & \cellcolor{gray!25}n/a\\

						\bottomrule
					\end{tabular}
				}
				
				\label{tab:table1}
		\end{table}

		\begin{table}[t]
			\caption{Evaluation on the \textbf{Compas} dataset using different models presented at the FairX. Bold indicates best result, and all the metrics score are higher as better. Synthetic Data Evaluation is only applicable to the Fair Generative Models (i.e.~TabFairGAN and Decaf).}
			\centering
				\resizebox{1.0\columnwidth}{!}{%
					\begin{tabular}{lccccccccc}\toprule
						& & \multicolumn{2}{c}{\textbf{Fairness Metrics}} &  \multicolumn{3}{c}{\textbf{Data Utility}} & \multicolumn{3}{c}{\textbf{Synthetic Data Evaluation}}\\
						\cmidrule(lr){3-4}  \cmidrule(lr){5-7} \cmidrule(lr){8-10}
						&\textbf{Protected} & DPR & EOR & ACC & AUC & F1- & $\alpha$- & $\beta$- & Authenticity\\ 
						& \textbf{Attribute} & & & & & Score & precision & recall &\\
						
						\midrule
						
						Correlation-  & Gender  & 0.43 $\pm$ .01 & 0.33 $\pm$ .01 & 0.64 $\pm$ .01 & 0.64 $\pm$ .01 & 0.59 $\pm$ .01 & \cellcolor{gray!25}n/a & \cellcolor{gray!25}n/a &  \cellcolor{gray!25}n/a\\
						Remover & Race  & 0.58 $\pm$ .01 & 0.63 $\pm$ .01 & 0.65 $\pm$ .01 & 0.64 $\pm$ .01 & 0.60 $\pm$ .01 &  \cellcolor{gray!25}n/a &  \cellcolor{gray!25}n/a &  \cellcolor{gray!25}n/a\\
						\midrule
						
						
						
						TabFairGAN & Gender  & 0.52 $\pm$ .01 & 0.42 $\pm$ .01 & \textbf{0.68 $\pm$ .01} &\textbf{ 0.68 $\pm$ .01} & \textbf{0.66 $\pm$ .01} & \textbf{0.84 $\pm$ .01} & \textbf{0.70 $\pm$ .01} & 0.37 $\pm$ .01\\
						& Race  & 0.50 $\pm$ .01 & 0.49 $\pm$ .01 & \textbf{0.69 $\pm$ .01} & \textbf{0.68 $\pm$ .01} & \textbf{0.64 $\pm$ .01} & \textbf{0.94 $\pm$ .01} & \textbf{0.75 $\pm$ .01} & 0.33 $\pm$ .01\\
						\midrule

						Decaf & Gender  & 0.87 $\pm$ .01 & 0.84 $\pm$ .01 & 0.45 $\pm$ .01 & 0.45 $\pm$ .01 & 0.42 $\pm$ .01 & 0.77 $\pm$ .01 & 0.45 $\pm$ .01 & \textbf{0.61 $\pm$ .01}\\
						& Race  & \textbf{0.99 $\pm$ .01} & \textbf{0.96 $\pm$ .01} & 0.45 $\pm$ .01 & 0.45 $\pm$ .01 & 0.42 $\pm$ .01 & 0.77 $\pm$ .01 & 0.45 $\pm$ .01 & \textbf{0.61 $\pm$ .01}\\
						
						\midrule
						
						FairDisco & Gender  & \textbf{0.97 $\pm$ .01} & 0.92 $\pm$ .01 & 0.55 $\pm$ .01 & 0.54 $\pm$ .01&  0.43 $\pm$ .01 &  \cellcolor{gray!25}n/a &  \cellcolor{gray!25}n/a & \cellcolor{gray!25}n/a \\
						& Race  & 0.87 $\pm$ .01 & 0.76 $\pm$ .01 & 0.53 $\pm$ .01 & 0.53 $\pm$ .01 & 0.44 $\pm$ .01 & \cellcolor{gray!25} n/a&  \cellcolor{gray!25}n/a & \cellcolor{gray!25}n/a \\
						\midrule
						
						
						Threshold  & Gender  & 0.92 $\pm$ .01 & \textbf{0.98 $\pm$ .01}& 0.65 $\pm$ .01 & 0.65 $\pm$ .01& 0.61 $\pm$ .01 & \cellcolor{gray!25}n/a &  \cellcolor{gray!25}n/a & \cellcolor{gray!25}n/a\\
						Optimizer& Race  & \textbf{0.99 $\pm$ .01} & 0.76 $\pm$ .01 & 0.63 $\pm$ .01 & 0.63 $\pm$ .01 & 0.60 $\pm$ .01&  \cellcolor{gray!25}n/a & \cellcolor{gray!25}n/a & \cellcolor{gray!25}n/a\\
						
						\midrule
						
						Original Data  & Gender  & 0.37 $\pm$ .01 & 0.28 $\pm$ .01 & 0.66 $\pm$ .01 & 0.65 $\pm$ .01 & 0.61 $\pm$ .01 &  \cellcolor{gray!25}n/a & \cellcolor{gray!25}n/a &  \cellcolor{gray!25}n/a\\
						& Race  & 0.54 $\pm$ .01 & 0.58 $\pm$ .01 & 0.66 $\pm$ .01 & 0.65 $\pm$ .01 & 0.61 $\pm$ .01 & \cellcolor{gray!25}n/a & \cellcolor{gray!25}n/a & \cellcolor{gray!25}n/a\\

						\bottomrule
					\end{tabular}
				}
				
				\label{tab:table2}
		\end{table}

		\section{Results and discussion} \label{sec:results}
		
		We now consider the fairness, data utility and synthetic data evaluation (only for in-processing generative models) of the models presented in this benchmarking tool. We also present the explainability analysis where we use the generated data by in-processing generative models and show how the fair generated data perform on down-streaming task and how the prediction is affected by the sensitive attributes. We also show the feature importance by using these explainability analysis.
		
		Table \ref{tab:table1} and \ref{tab:table2} shows the performance of the bias mitigation algorithms for the \textit{Adult-Income} dataset and \textit{Compas} dataset respectively. We run experiment using different Protected attributes \footnote{For the sake of brevity, we could not include additional results using other datasets---we refer the reader to the FairX repository for these results. Some metrics like precision, recall, fairness through unawareness (FTU), and plots like fairness-accuracy trade-offs were similarly omitted.}. Besides, fairness and data utility, we add synthetic data evaluation for the output of TabFairGAN \footnote{\url{https://github.com/amirarsalan90/TabFairGAN}}, and Decaf \footnote{\url{https://github.com/vanderschaarlab/synthcity}}.
		
		\begin{figure}[t]
			\centering
			\includegraphics[width = 0.7\textwidth]{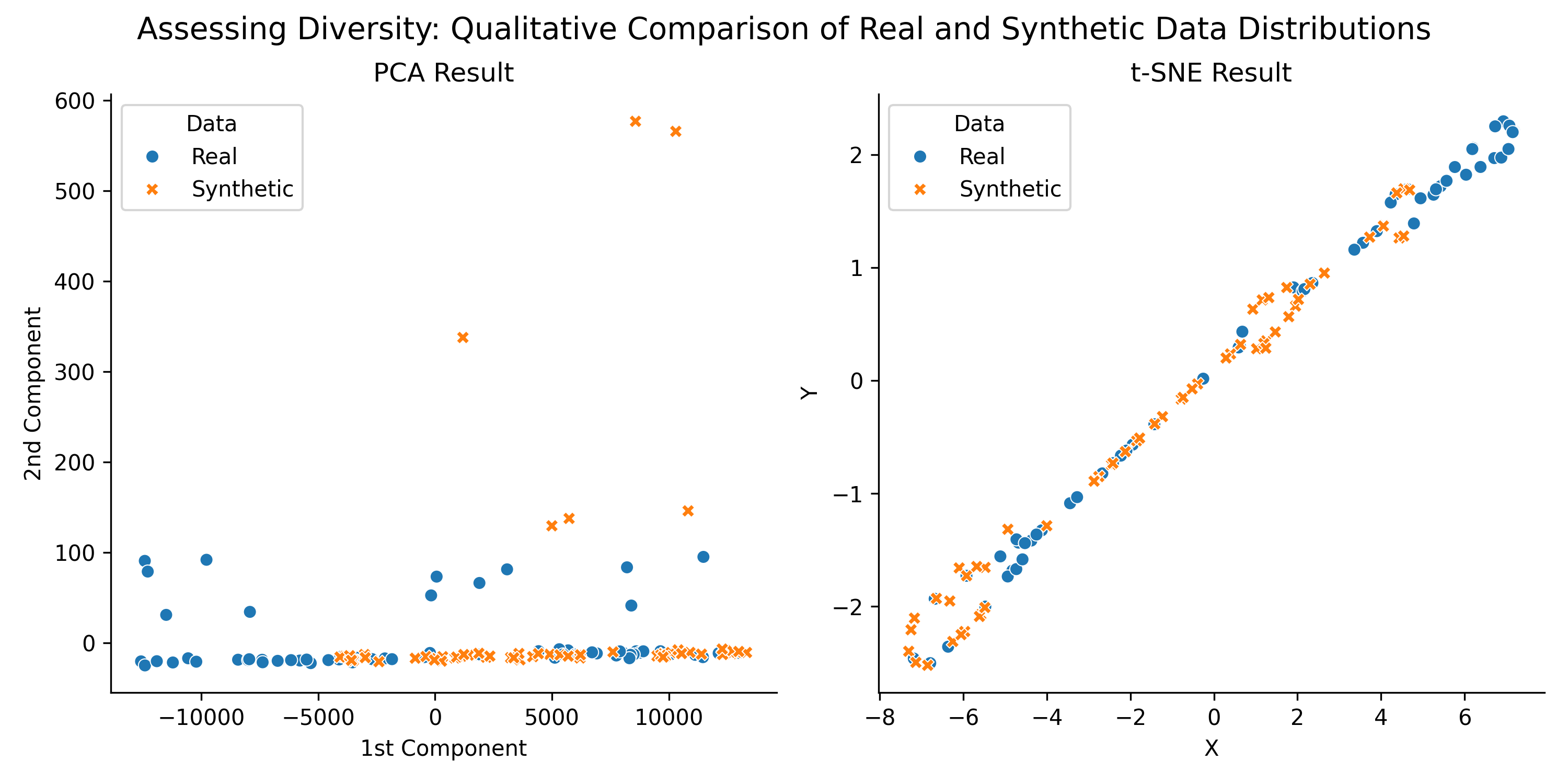}
			\caption{PCA and t-SNE plots of the original data and Synthetic data generated by TabFairGAN. Here each dot represents a record, if the generative model learns the original data distribution then the dots should overlap with each other. Dataset: `Adult-Income', Protective attribute: `sex'.}
			\label{fig:tsne}
		\end{figure}
		
		From the table, we see for the generative fair models, TabFairGAN is performing well comparing with the Decaf in both datasets with both protected attributes. The $\alpha-$precision, $\beta-$recall scores of TabFairGAN is better than Decaf, this represents the synthetic data quality of TabFairGAN is superior than Decaf. On the other hand, TabFairGAN perform poorly in the fairness evaluation for the `race' protected attribute of the \textit{Adult-Income} dataset. Whereas In-processing technique FairDisco \footnote{\url{https://github.com/SoftWiser-group/FairDisCo}} performs well in terms of fairness and data utility.
		
		On the visual evaluation of fair synthetic data, we use the synthetic data generated by TabFairGAN. Figure \ref{fig:tsne} shows the PCA and t-SNE plots of the synthetic data generated by TabFairGAN. We show how closely the synthetic data distribution is matching with the original data. If the generative model can capture the original data distribution, original and synthetic data should overlap with each other on the PCA and t-SNE plot. Figure \ref{fig:tsne} shows that data generated by TabFairGAN partially learned the data distribution of the original data.
		
		\begin{figure}[t]
			\centering
			\includegraphics[width = 0.45\textwidth]{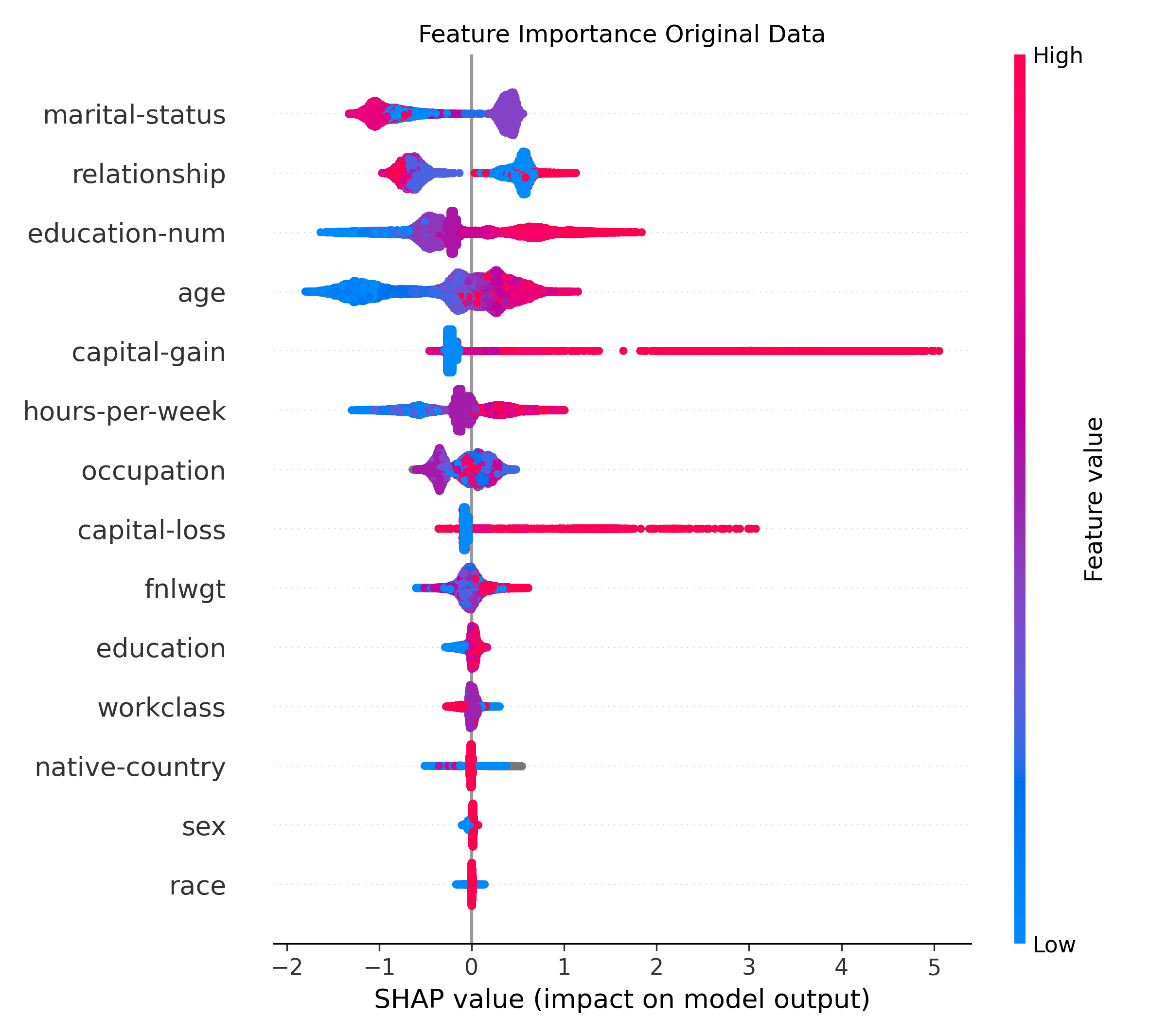}
			\includegraphics[width = 0.45\textwidth]{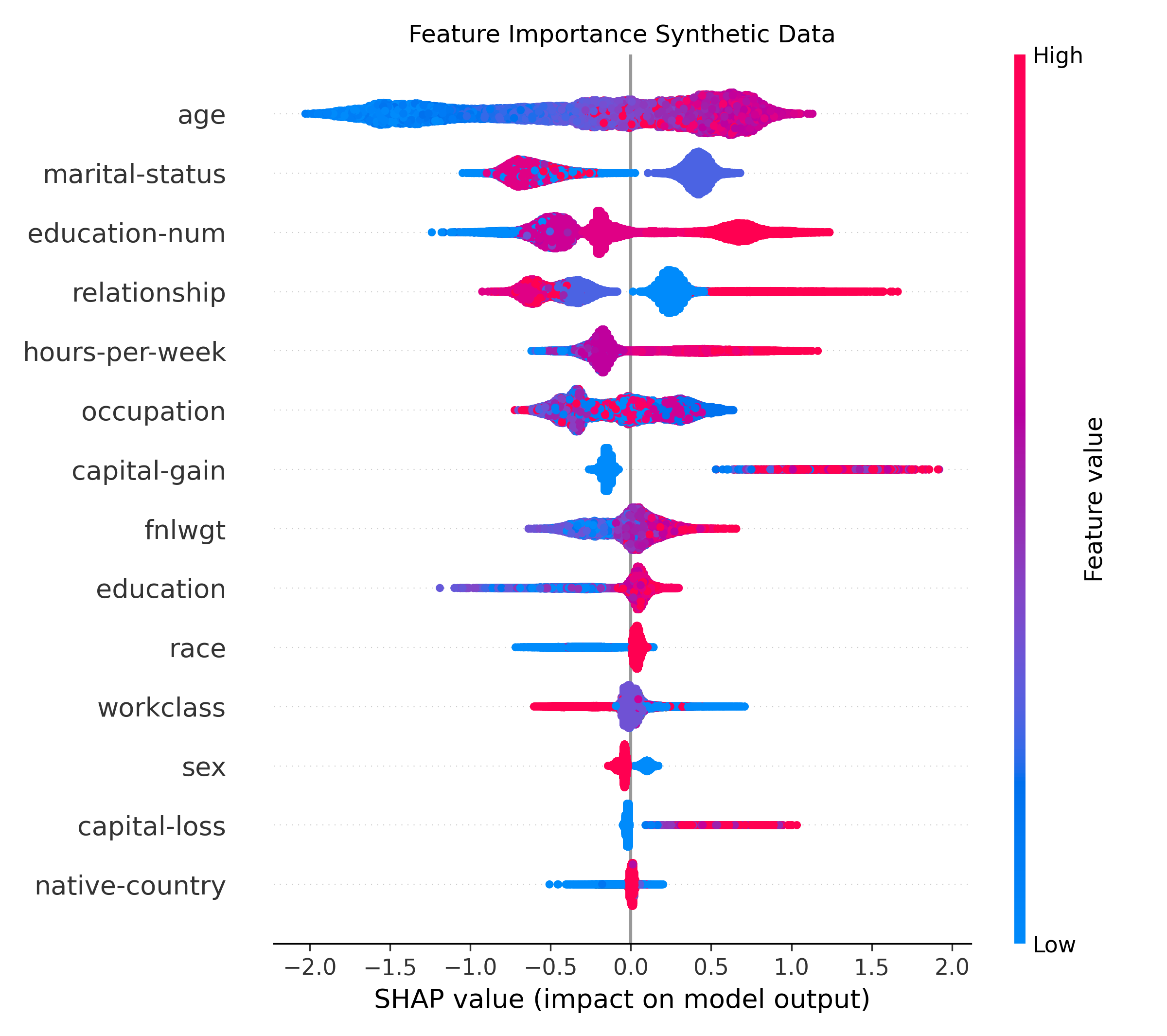}
			\caption{Feature Importance on Prediction task on the Original Data (left) and Synthetic Data (right) by TabFairGAN, the Sensitive Feature here is `sex', The Feature Value of Sensitive Attribute in Synthetic Data is less than Original Data.}
			\label{fig:explain}
		\end{figure}
		
		In Figure \ref{fig:explain}, we show the feature importance for a down-streaming task to predict the target attribute of the \textit{Adult-Income} dataset where the `Sensitive attribute' is `sex'. We compared the feature importance of original data with the synthetic data generated by TabFairGAN. We can see the feature importance of the synthetic data is lower than the original data. This means the synthetic data generated by the TabFairGAN is less biased towards entity.
		
		\begin{figure}[t]
			\centering
			\includegraphics[width = \textwidth]{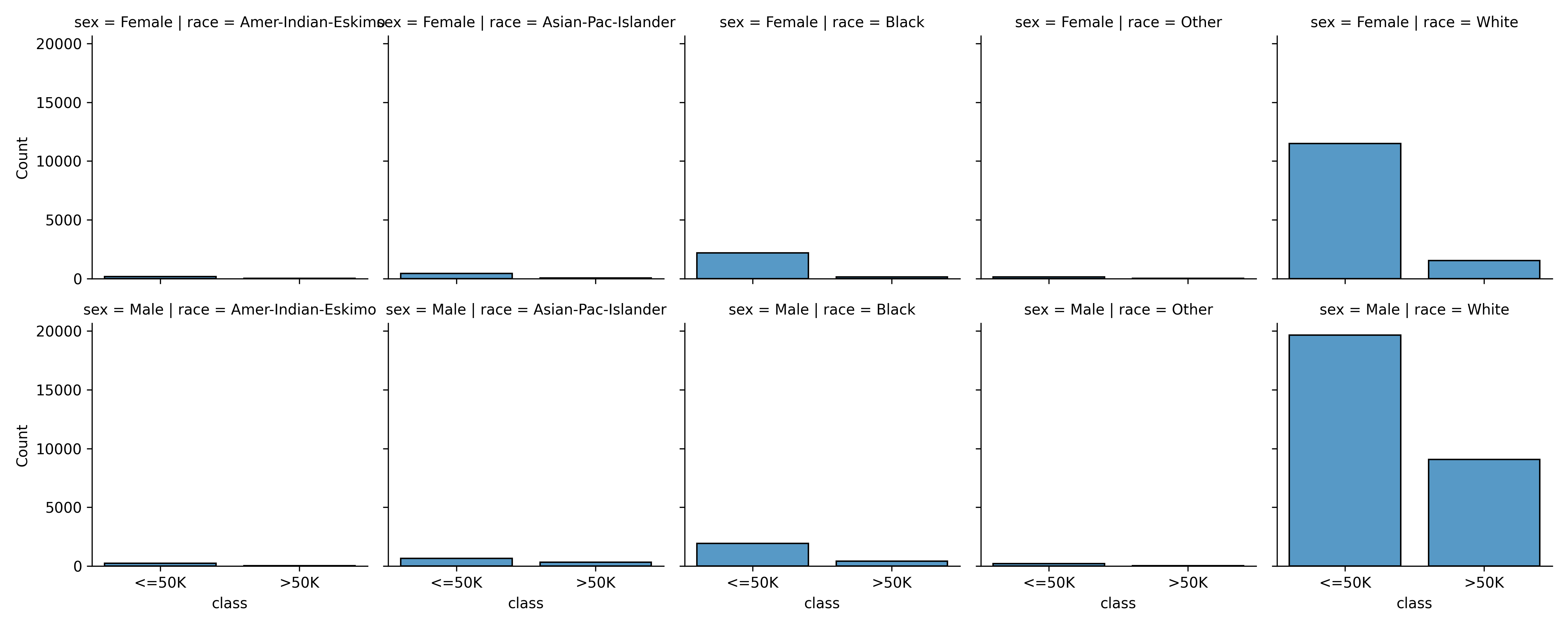}
			\caption{Representation of `sex' and `race' features on the target class, here we can see the dataset is heavily in favor of white people.}
			\label{fig:ib}
		\end{figure}
		
		Finally, Figure \ref{fig:ib} shows the intersectional bias on the \textit{Adult-Income} dataset. We plot the percentage of `salary-income' for both `race' and `sex' protected attributes. We see in the dataset, decisions are given in favor towards white people.

		\section{Conclusion and future work} \label{sec:conclusion}
		
		Massive amounts of data are being produced everyday. Unfortunately, much of this data contains human or machine biases. Furthermore, the usage of recommendation system has increased with advancements in artificial intelligence. But if we use biased data to train a recommendation system, there is a high chance that the recommendation system will yield unfair decision towards some demographics. To mitigate this issue, researchers have developed various measure to mitigate the bias from the dataset, or to train the model in such a way that the model produces bias-free data. To help in this process, benchmarking tools equipped with different bias-mitigation techniques and evaluation metrics were developed over the years. But these benchmarking tools commonly lack the option to evaluate generative models or to train them. We therefore presented FairX, an open-source, modular, fairness benchmarking tool. FairX comes with a data-loader, supports model training, and has an evaluation module. FairX provides support for training fair generative models and for evaluating the synthetic data created by them. FairX also contains various fairness evaluation metrics, data utility evaluation metrics and different plotting techniques to help users to evaluate models and visualize outcomes. FairX comes with support for explainability analysis of a prediction using the dataset (both original and synthetic) and shows feature importance. 
		We believe FairX will help the researchers and mitigate the gap of not having fair generative models and way of evaluating synthetic data.
		
		In the future, we intend to extend FairX to be able to handle other modalities in addition to tabular and image data, for example text and video.
		Also, we will add wider range of evaluation metrics for both synthetic data utility and fairness metrics. 
		For the models, we plan to add text based and more tabular and image based fair generative models \cite{li2022fairgan, ramachandranpillai2023fair, choi2020fair, ramachandranpillai2024bt}.
		In this version of FairX, we do not have option to add custom models, but we plan to add this features in future version, so users can use their own model and use all the functionalities of FairX for their model. We also plan to add hyper-parameter optimization feature for the models so, we can find the optimal parameters and best result.
		Finally, we plan to add functionalities to evaluate the output of large language models.

		\begin{acknowledgments}
			The work was partially funded by the Knut and Alice Wallenberg Foundation, and the TAILOR Network of Excellence for trustworthy AI (EC Grant Agreement 952215).
			Portions of this work were carried out using the AIOps/Stellar facilities funded by the Excellence Center at Linköping–Lund in Information Technology (ELLIIT).
		\end{acknowledgments}
		
		\bibliography{sample-1col}
		
		\appendix
		
		\section{Detailed Usage} 
		\label{appsec:usage}
		
		In this section, we present different sample code example of our tool. We give a brief description of each module and their corresponding class description and function details.
		
		\paragraph{Dataset usage.} \label{sub:dataset-usage}
		To use the dataset already pre-loaded with the tool, we need to use the \textsc{BaseDataClass}. This class takes three hyperparameters as input; \textsc{dataset\_name}, \textsc{sensitive\_attirbute} and a boolean flag for attaching the target variable with the main dataframe. \textsc{BaseDataClass} has two functions, \textsc{preprocess\_data()} and \textsc{split\_data()} to preprocess the dataset using categorical, numerical transformation and split the dataset for training and testing purpose respectively. 
		
		\begin{lstlisting}[language=Python, caption=Using \textsc{BaseDataset} Class.]
			from fairX.dataset import BaseDataClass
			
			dataset_name = 'Adult-Income'
			sensitive_attribute = 'race'
			attach_target = True
			data_module = BaseDataClass(dataset_name, sensitive_attribute, attach_target)
		\end{lstlisting}

		\paragraph{Model usage.}
		\label{sub:model-usage}
		
		We add three kinds of bias-removal techniques under the models folder of FairX. The list of available models can be found in Table 2. Here is an example usage of in-processing algorithm called \textit{TabFairGAN}. After initializing the Model, we train the it by calling the \textsc{fit()} function which takes the dataset, batch size and number of epochs as parameters. After training, for the fair generative models (TabFairGAN and Decaf), synthetic data will be automatically saved in the working directory.
		
		\begin{lstlisting}[language=Python, caption=Using Models.]
			from fairX.models.inprocessing import TabFairGAN
			
			data_module = BaseDataClass(dataset_name, sensitive_attribute, attach_target)
			under_prev = 'Female'
			y_desire = '>50K'
			tabfairgan = TabFairGAN(under_prev, y_desire)
			tabfairgan.fit(data_module, batch_size = 256, epochs = 1000)
		\end{lstlisting}

		\paragraph{Metrics usage.}
		\label{sub:metrics-usage}
		
		Here, we give a sample code for measuring the fairness and data utilities with a dataset that is already part of the FairX system. Both \textsc{FairnessUtils} and \textsc{DataUtilsMetrics} class takes the dataset as input and then we call the \textsc{evaluate\_fairness()} and \textsc{evaluate\_utility()} function to measure the fairness data utilities respectively. The result is stored as a dictionary file.
		
		\begin{lstlisting}[language=Python, caption=Using Fairness \& Data utility Metrics.]
			from fairX.metrics import FairnessUtils
			from fairX.metrics import DataUtilsMetrics
			from fairX.dataset import BaseDataClass
			
			data_module = BaseDataClass(dataset_name, sensitive_attribute, attach_target)
			cat_transformer, num_scaler, transformed_data = data_module.preprocess_data()
			splitted_data = data_module.split_data(transformed_data)
			fairness_measurement = FairnessUtils(splitted_data)
			utility_measurement = DataUtilsMetrics(splitted_data)
			fairness_res = fairness_measurement.evaluate_fairness()
			datautils_res = utility_measurement.evaluate_utility()
			print(fairness_res)
			print(datautils_res)
		\end{lstlisting}
		
		The following code example is to use the \textsc{CustomDataClass} to load custom dataset in FairX. We need to give the dataset path, list of sensitive attributes and a boolean operator for attaching the target. This code also shows the usage of synthetic data evaluation using the \textsc{SyntheticEvaluation} class.
		
		\begin{lstlisting}[language=Python, caption=Using Synthetic Data Evaluation Metrics with Custom Data Loader.]
			from fairX.metrics import SyntheticEvaluation
			from fairX.dataset import BaseDataClass
			from fairX.dataset import CustomDataClass
			
			original_data = BaseDataClass(dataset_name, sensitive_attribute, attach_target)
			generated_data = CustomDataClass(generated_data_path, sensitive_attribute, attach_target)
			synthetic_evaluation_class = SyntheticEvaluation(original_data, generated_data)
			synthetic_data_measurement = synthetic_evaluation_class.evaluate_synthetic()
			print(synthetic_data_measurement)
		\end{lstlisting}


		
	\end{document}